\pdfoutput=1

\documentclass[11pt]{article}

\usepackage{multirow}
\usepackage{multicol}
\usepackage{amssymb}
\usepackage{subcaption}
\usepackage{bm}
\usepackage{cite}
\usepackage{booktabs}
\usepackage{graphics}
\usepackage{amsmath,epsfig}

\usepackage{bbding}
\usepackage{amssymb}
\usepackage{pifont}
\newcommand{\cmark}{\ding{51}}%
\newcommand{\xmark}{\ding{55}}%

\usepackage{xcolor}

\usepackage{algorithm}
\usepackage[noend]{algorithmic}
\usepackage{eqparbox}

\definecolor{visualorange}{HTML}{F4B183}
\definecolor{textgreen}{HTML}{70AD47}
\definecolor{applegreen}{rgb}{0.55, 0.71, 0.0}

\usepackage{makecell}
\usepackage{fancybox}
\usepackage[many]{tcolorbox}

\usepackage{EMNLP2022}
\usepackage{times}
\usepackage{latexsym}

\usepackage[T1]{fontenc}

\usepackage[utf8]{inputenc}

\usepackage{microtype}

\usepackage{inconsolata}

\title{MoSE: Modality Split and Ensemble for \\ Multimodal Knowledge Graph Completion}

\author{
Yu Zhao\textsuperscript{1}\;\;\;Xiangrui Cai\textsuperscript{1}\thanks{\; Corresponding author.}\;\;\;Yike Wu\textsuperscript{2} \;\;\;Haiwei Zhang\textsuperscript{1}  \\ {\bf Ying Zhang\textsuperscript{3}\;\;\;Guoqing Zhao\textsuperscript{4}\;\;\;Ning Jiang\textsuperscript{4}} \\
\textsuperscript{1} College of Cyber Science, TKLNDST, Nankai University, Tianjin, China \; \\
\textsuperscript{2} School of Journalism and Communication, Nankai University, Tianjin, China \; \\
\textsuperscript{3} College of Computer Science, Nankai University, Tianjin, China \; \\
\textsuperscript{4} Mashang Consumer Finance Co, Ltd \\
{\tt zhaoyu@dbis.nankai.edu.cn}, {\tt \{caixr,wuyike,zhhaiwei,yingzhang\}@nankai.edu.cn}
  }

\begin{document}
\maketitle
\begin{abstract}
Multimodal knowledge graph completion (MKGC) aims to predict missing entities in MKGs.
Previous works usually share relation representation across modalities. This results in mutual interference between modalities during training, since for a pair of entities, the relation from one modality probably contradicts that from another modality. 
Furthermore, making a unified prediction based on the shared relation representation treats the input in different modalities equally, while their importance to the MKGC task should be different. 
In this paper, we propose \texttt{\textbf{MoSE}}, a \textbf{Mo}dality \textbf{S}plit representation learning and \textbf{E}nsemble inference framework for MKGC.
Specifically, in the training phase, we learn modality-split relation embeddings for each modality instead of a single modality-shared one, which alleviates the modality interference. Based on these embeddings, in the inference phase, we first make modality-split predictions and then exploit various ensemble methods to combine the predictions with different weights, which models the modality importance dynamically. 
Experimental results on three KG datasets show that \texttt{MoSE} outperforms state-of-the-art MKGC methods. Codes are available at \href{https://github.com/OreOZhao/MoSE4MKGC}{https://github.com/OreOZhao/MoSE4MKGC}.

\end{abstract}

\section{Introduction}

\begin{figure}[t]
     \centering
     \begin{subfigure}[b]{0.48\textwidth}
         \centering
         \includegraphics[width=\textwidth]{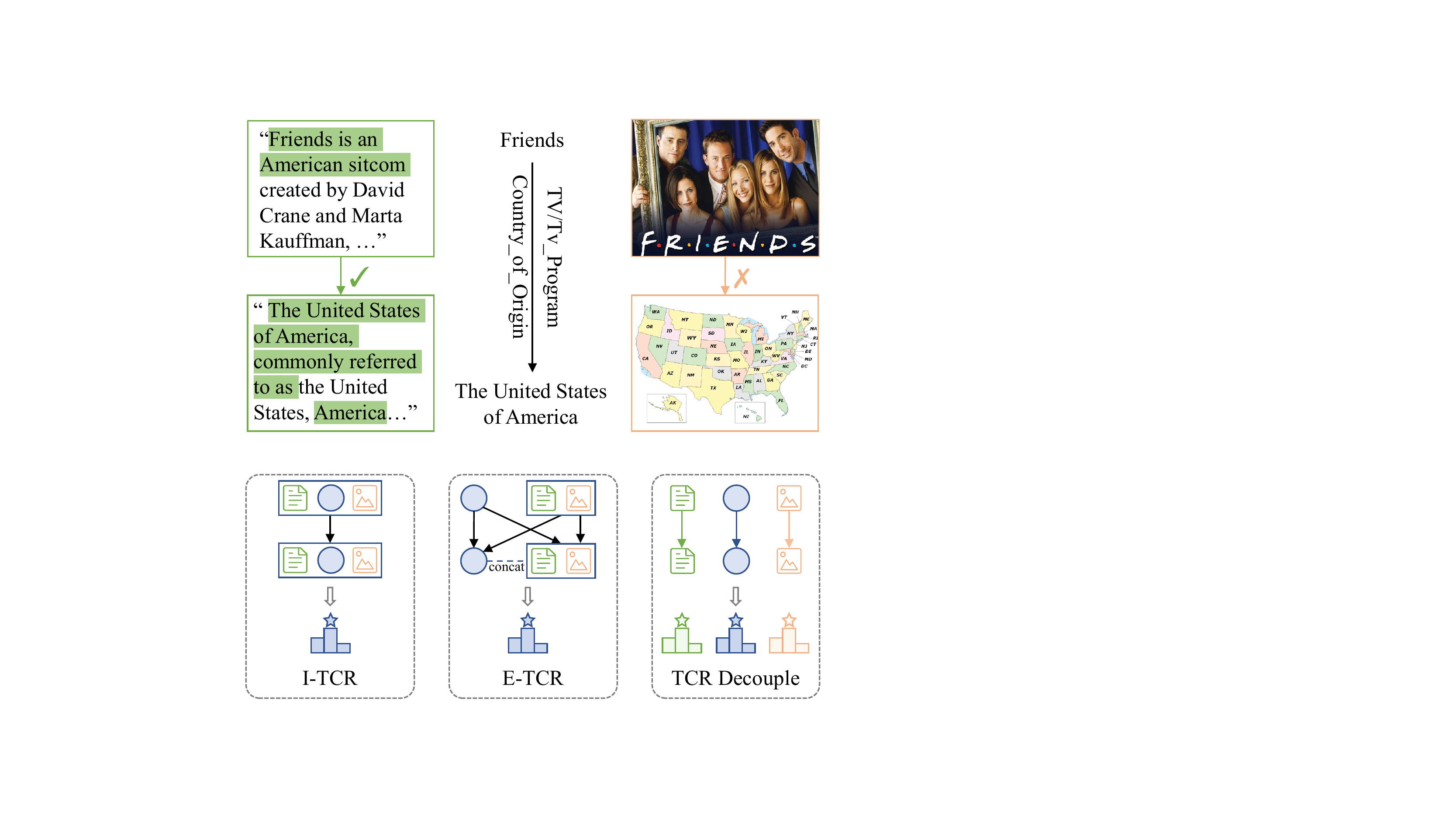}
         \caption{An example of contradictory relations between modalities.}
         \label{fig:intro_example}
     \end{subfigure}
    \vspace*{\fill}
     \begin{subfigure}[b]{0.48\textwidth}
         \centering
         \includegraphics[width=\textwidth]{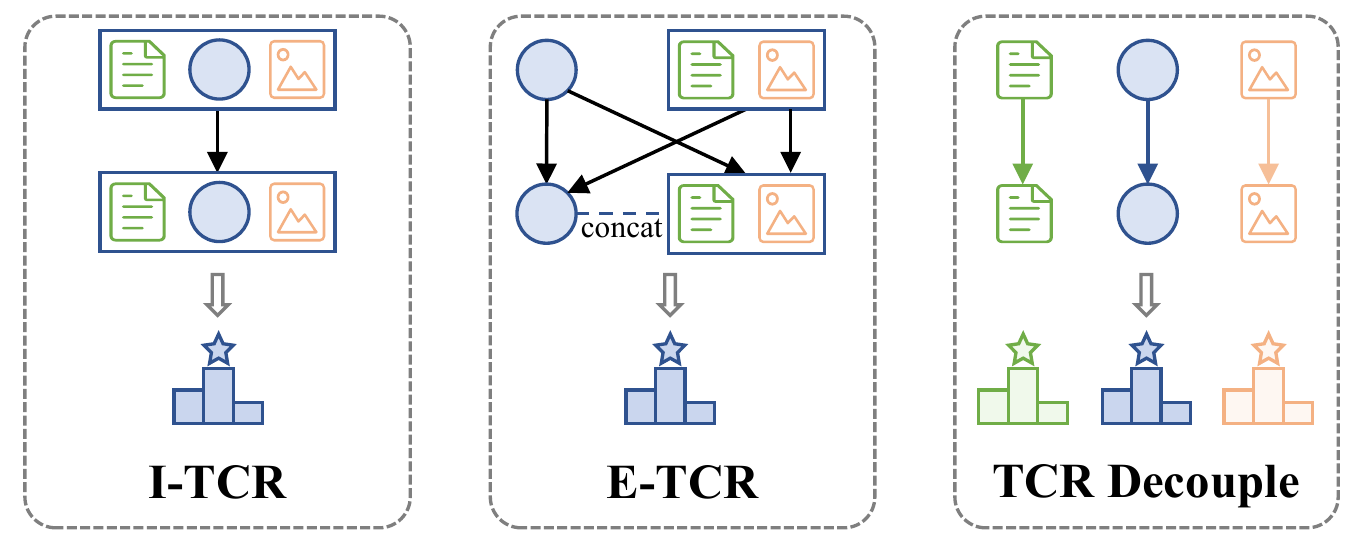}
         \caption{Implicit (I-) and Explicit (E-) Tight-Coupling Relation (TCR), and TCR Decouple (Ours). Note that the arrows with the same color represent \textit{the same} relation embedding.}
         \label{fig:intro_TCR}
     \end{subfigure}
     \caption{(a) displays an example of multimodal triples, in which text modality demonstrates relevant relation \textcolor{textgreen}{\cmark} and visual modality demonstrates irrelevant relation \textcolor{visualorange}{\xmark}. (b) visualizes existing approaches with implicit and explicit tight-coupling relation (TCR), and our approach with decoupled relations.}
\end{figure}
Multimodal knowledge graphs (MKGs) organize multimodal facts in the form of entities and relations, and have been successfully applied to various knowledge-driven tasks \cite{zhu2022mmkgsurvey,zhang2018mner,marino2019okvqa,sun2020mmkgrs}.
To address the inherent incomplete problems in MKGs, 
multimodal knowledge graph completion (MKGC) has been proposed \cite{xie2016DKRL,xie2017IKRL}, which utilizes auxiliary visual or text information to help predict missing entities.
As shown in Figure \ref{fig:intro_example}, given the head entity \ovalbox{Friends} and the relation $\stackrel{country}{\longrightarrow}$, MKGC is required to predict the tail entity \ovalbox{The United States of America}. It can be observed that the descriptions attached to entities provide supplementary information for entity prediction.

Existing MKGC methods usually share a common relation embedding across all modalities for a pair of entities, which tightly couples multiple relations from different modalities. We define this paradigm of MKGC as \textbf{Tight-Coupling Relation (TCR)}.
As shown in Figure \ref{fig:intro_TCR}, according to the way that the relations from different modalities are coupled, existing methods can be roughly divided into two categories: Implicit TCR (I-TCR) methods and Explicit TCR (E-TCR) methods. I-TCR methods \cite{wang2019transAE,wang2021rsme} usually first fuse multimodal information of an entity into a single embedding, and then learn a unified relation representation based on the embedding.
E-TCR methods \cite{xie2016DKRL,xie2017IKRL,mousselly2018MKB} directly model the relationship between separate multimodal information of entities without fusion. They usually learn a
single relation embedding to simultaneously represent all intra-modal and inter-modal relations.

Although existing MKGC methods have achieved promising results, they are limited 
by TCR in two folds:
(1) \textbf{Modality relation contradiction.} The TCR usually simultaneously represents multiple relations from different modalities only with a single embedding. However, for a pair of entities, the relation from one modality probably contradicts that from another modality. For example, as shown in Figure \ref{fig:intro_example},
the description \textit{"American sitcom"} of entity \ovalbox{Friends} demonstrates the relation $\stackrel{country}{\longrightarrow}$ to entity \ovalbox{The United States of America}, while the images do not.
The inherent contradiction of TCR results in modality interference during representation learning in MKGs.
(2) \textbf{Modality difference ignorance.}
Based on TCR, existing methods usually treat the input in different modalities equally and make a unified prediction, which ignores the difference of modality importance. 
However, different modalities vary in data quality and entity coverage, and should contribute to the final prediction in varying degrees. 

To overcome the above limitations, we propose a \textbf{Mo}dality \textbf{S}plit learning and \textbf{E}nsemble inference framework, \texttt{MoSE}.
As shown in Figure \ref{fig:intro_TCR}, in the training phase, \texttt{MoSE} decouples TCR and learns multiple modality-split relation embeddings instead of a single modality-shared one, which alleviates mutual interference between modalities.
In the inference phase, \texttt{MoSE} first makes predictions for each modality separately based on the modality-split embeddings, and then merges them into the final prediction. 
We explore the best combination of modality predictions with various ensemble methods, and model the modality importance by modulating the modality weights dynamically.
Experimental results and analysis on three widely-used datasets show that \texttt{MoSE} outperforms state-of-the-art methods for MKGC task. 

Overall, the contributions of this paper can be summarized as follows:
\begin{itemize}
    \item To the best of our knowledge, we are the first to deal with the modality contradiction of relation representation and discuss modality importance in MKGC task.
    \item We propose a modality-split learning and ensemble inference framework \texttt{MoSE} for MKGC, which decouples the tight-coupling relation embedding into modality-split ones in the training phase, and modulate modality importance adaptively in the inference phase.
    \item Experiment results on three datasets demonstrate that \texttt{MoSE} outperforms 9 baselines and obtain the state-of-the-art performance in MKGC task. The results also show that text modality is a useful complement for MKGC rather than visual modality.
\end{itemize}

\section{Related Work}

Existing researches on MKGC mainly focus on extending unimodal knowledge graph embedding (KGE) models to further exploit multimodal information.
We notice that for a pair of entities in an MKG, existing multimodal KGE methods all exploit a modality-shared relation embedding which tightly couples multiple relations from different modalities, which we call \textbf{Tight-Coupling Relation (TCR)}. 
We divide existing methods to two categories: implicit tight-coupling relation (I-TCR) methods and explicit tight-coupling relation (E-TCR) methods.

\subsection{Implicit TCR Methods}
I-TCR methods \cite{wang2019transAE,wang2021rsme} fuse multiple modalities into a unified entity embedding and utilize a shared relation representation as shown in Figure \ref{fig:intro_TCR}. 
Thus the learned relation implicitly fuses multimodal relations.
TransAE \cite{wang2019transAE} extends TransE with auto-encoder fusing visual and text information into entity representation.
Recently, RSME \cite{wang2021rsme} notices the noise in visual modality and propose a forget gate to adjust the fusion rate of image to entity embeddings and reaches state-of-the-art (SOTA) performance.
Though I-TCR methods show promising improvements, they neglect modality contradictions in modality-shared relation representation.
Moreover, they make unified predictions without assessing whether the modality information is relevant to final predictions.
In terms of SOTA RSME, the fusion ratio of visual information is determined by image information itself, i.e. similarity, regardless of modality importance to final prediction.

\subsection{Explicit TCR Methods}
E-TCR methods \cite{xie2016DKRL,xie2017IKRL,mousselly2018MKB} utilize a shared relation embedding which tightly couples multiple relations between intra-modal and inter-modal entities.
E-TCR methods learn representations with an overall score across all modalities: structure-structure, structure-visual/text, visual/text-structure, visual/text-visual/text, all connected by a single modality-shared relation embedding as shown in Figure \ref{fig:intro_TCR}, which explicitly tightly couples multiple relations.
DKRL \cite{xie2016DKRL} and IKRL \cite{xie2017IKRL} extend TransE with text and visual modality respectively. 
MKB \cite{mousselly2018MKB} extends IKRL \cite{xie2017IKRL} from visual modality to visual-text multi-modalities.
Although E-TCR methods project multimodal features to a common latent space, the inherent semantic contradiction of relations between different modalities is not eliminated. 
Moreover, they utilize weighted concatenation of multimodal entities to make a unified prediction and does not consider modality importance either.

\section{Methodology}

\begin{figure*}[!t]
    \centering
    \includegraphics[width=1\textwidth]{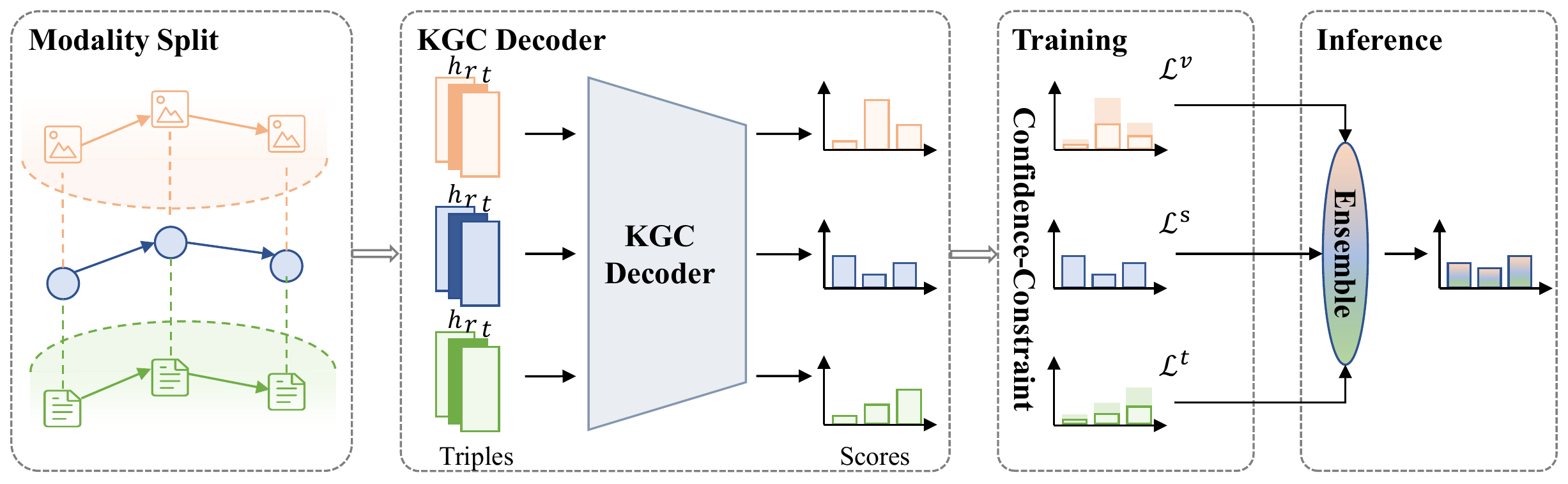}
    \caption{The framework of \textbf{Mo}dality-\textbf{S}plit learning and \textbf{E}nsemble Inference, \texttt{MoSE}, for multimodal knowledge graph completion.}
    \label{fig: model_overview}
\end{figure*}

\subsection{Preliminaries}
In this section, we introduce the notation used in this paper and formulate the MKGC task.

\textbf{KGC task.}
Knowledge graph is a collection of factual triples $\mathcal{G} = \{(h,r,t)\}$, where head entity and tail entity $h,t \in \mathcal{E}$ and relation $r \in \mathcal{R}$. 
The KGE model (1) represents entities and relations to vectors $\boldsymbol{h},\boldsymbol{r}, \boldsymbol{t}$, (2) utilizes a score function $f(\boldsymbol{h},\boldsymbol{r},\boldsymbol{t}): \mathcal{E} \times \mathcal{R} \times \mathcal{E} \rightarrow \mathbb{R}$ to decode the plausibility of a triple to scores.
For a particular query $q = (h,r,?)$, the KGC task aims at ranking all possible entities and obtaining prediction preference.

\textbf{MKGC task. }In MKGs, each entity $e \in \mathcal{E}$ has multimodal embeddings $\boldsymbol{e}_m, m\in\mathcal{M}=\{\mathcal{S}, \mathcal{V},\mathcal{T}\}$, which denotes structure, visual, and text modality respectively. 
We use $\boldsymbol{e}_s, \boldsymbol{e}_v, \boldsymbol{e}_t$ to denote corresponding entity embedding, where visual and text embedding is projection of extracted features $\boldsymbol{e}_v = W_v f_v, \boldsymbol{e}_t = W_t f_t$.

\textbf{TCR methods.}
I-TCR methods design a fusion mechanism $\Phi(\{ \boldsymbol{e}_m\}), m\in \mathcal{M}$ to get a fused embedding of multimodal entities and extends the score function as $f(h,r,t)=f(\Phi(\{ \boldsymbol{h}_m \}),\boldsymbol{r},\Phi(\{ \boldsymbol{t}_m \})$.
E-TCR methods utilize score function across all modalities, where $f(h,r,t)=\sum_{i=1}^{|\mathcal{M}|} \sum_{j=1}^{|\mathcal{M}|} f(\boldsymbol{h}_i,\boldsymbol{r},\boldsymbol{t}_j)$.
It is worth noting that both ways utilize a modality-shared relation representation.

\subsection{Overview}
Figure \ref{fig: model_overview} shows our \textbf{Mo}dality \textbf{S}plit learning and \textbf{E}nsemble inference framework, \texttt{MoSE}, for multimodal knowledge graph completion task.
We first decouple TCR to modality-split relation embeddings corresponding to each modality.
With the decoupled TCR, 
we construct modality-split triple representations for each modality to prevent modality interference in representations.
Through the KGE score function, the modality-split representations are decoded to corresponding score distributions.
In the training phase, we train modality-split entity and relation representations with intra-modal scores simultaneously.
Considering visual and text modalities usually embody more contradictory and uncertain noise than structure modality, we apply confidence-constraint training objectives for the two modalities.
In the inference phase, we exploit ensemble inference to combine the modality-split predictions and obtain the final predictions.
We explored three kinds of ensemble inference methods aiming at modeling modality importance.

\subsection{Modality-Split MKG Construction}

In our paper, we assume that the TCR embedding used in existing methods represents multiple contradictory relations simultaneously and results in modality interference.
Thus we propose to decouple the TCR and construct a modality-split MKG.
With decoupling the contradictory relations from a modality-shared embedding to multiple modality-split relation embeddings, \texttt{MoSE} alleviates modality interference in relation representations.
Formally, we construct decoupled modality-split relation embeddings for each relation type.
In this paper, we construct structure, visual and text modality relation embedding $\boldsymbol{r}_s$ $\boldsymbol{r}_v$ and $\boldsymbol{r}_t$ for relation $r$. 
Since relation representation is decoupled, we also avoid modality fusion in entities, which also prevents interference within entity representations.
Together with the modality-split entity and relations, we form a modality-split KG of multiple unimodal KGs which have identical topology but different entity and relation representations.

\subsection{KGC Decoder}
With the modality-split KG construction, the score function is also separated to multiple scores denoted as
$f_m(h,r,t) = f(\boldsymbol{h}_m,\boldsymbol{r}_m,\boldsymbol{t}_m), m \in \mathcal{M}$.
With the modality-split architecture, \texttt{MoSE} is able to present score distribution for each modality.
For each query triple $(h,r,t)$, the decoder gives different scores depending on the learned representation, which intuitively reflects the strengths and limitations of each modality for entity predictions.

\subsection{Training}
We utilize multi-class cross-entropy (CE) loss for training following \newcite{lacroix2018fbkbc}. 
Given a query $(h,r,?)$, we construct corrupted triples by replacing the tail entity with randomly selected entities in $\mathcal{E}$.
We also construct reverse triple $(?, r^{-1}, h)$ for each triple in the training set and apply the same setting.
For all triples, KGC decoder provides corresponding probability of truth $p_m(t|(h,r))=softmax(f_m(h,r,t))$
computed with a $softmax$ applied to the output of the score function. 
We denote $p_m(t|(h,r))$ as probability obtained in modality $m$.
The CE loss of modality $m$ can be calculated as Equation \eqref{eq:ce}. 
\begin{equation} \label{eq:ce}
\begin{aligned}
    \mathcal{L}_m=& - \sum_t^{|\mathcal{E}|} log(p_m(t|(h,r))\\
    =&CE(p_m(t|(h,r))) \,.
\end{aligned}
\end{equation}

\textbf{Confidence-constraint Training.}
Visual and text modalities usually embody contradictory information due to data complexity and diversity. 
We notice that the contradiction lies in the fact that the modality information of \textit{entity} is not always relevant to the \textit{knowledge} of factual triple.
In consequence, visual and text modality usually present uncertainty.
To ease the uncertainty, we train the visual or text modality KGC in a confidence-constraint manner with a temperature-scaling technique \cite{guo2017tempscaling}.
Since the predicted probability can approximately represent the confidence score of predictions, we simply compact the probability distribution by adding a temperature parameter $\mathcal{T}$ to the output of the KGC decoder as Equation \eqref{eq: cc} and extend CE loss to a confidence-constraint form as Equation \eqref{eq:cc ce}. 
In this way, the confidence of predictions is constrained and the distribution is softened while prediction results remain the same.
Formally, we use the confidence-constraint loss for visual and text modality as $\mathcal{L}_v^{cc}$ and $\mathcal{L}_t^{cc}$ respectively.
\begin{equation} \label{eq: cc}
    p^{cc}_m(h,r,t) = softmax(\frac{f_m(h,r,t)}{\mathcal{T}})\,.
\end{equation}
\begin{equation} \label{eq:cc ce}
    \mathcal{L}_m^{cc}=CE(p_m^{cc}(t|(h,r)))\,.
\end{equation}

\textbf{Overall Objective.}
In the training phase, we simultaneously train three modalities to learn intra-modal representations separately after the overall objective $\mathcal{L}_{KGC}$. 
The overall KGC objective is the sum of modality losses as Equation \eqref{eq:overall loss}.

\begin{equation} \label{eq:overall loss}
    \mathcal{L}_{KGC} = \mathcal{L}_s + \mathcal{L}_v^{cc} + \mathcal{L}_t^{cc} \,.
\end{equation}

\subsection{Inference}
In the inference phase, we explore combination mechanisms of modality predictions by modeling modality weights.
Modalities have strengths and limitations due to data quality and entity coverage, which are always complementary to each other.
Appropriately adjusting modality weights to fully exploit the complementary strengths would lead to better prediction performance.

\textbf{Ensemble Inference.}
Inspired by \newcite{chen2020multilingual}, we exploit ensemble inference to obtain final predictions.
We propose to directly combine scores instead of ranks since information from score distributions may get lost during the ranking process.
For each query, we obtain three score distributions $f_m(h,r,t), m \in \mathcal{M} = \{\mathcal{S}, \mathcal{V}, \mathcal{T}\}$ from three modalities, which could directly reflect the strengths and limitations of modality for entity predictions.
The scores are combined as Equation \eqref{eq:ensemble inference}.
\begin{equation} \label{eq:ensemble inference}
    \mathcal{F}(h,r,t) = \sum_{m \in \mathcal{M}} w_m f_m(h,r,t) \,.
\end{equation} 

Next, we propose three variants of \texttt{MoSE}: \texttt{MoSE-AI}, \texttt{MoSE-BI} and \texttt{MoSE-MI}, which varies in the modality weight $w_m$ calculation. 
We utilize a small amount of unbiased meta-set to learn modality weights that can be finely transferred to test-set. We choose validation-set as meta-set.

\textbf{Equal-importance Average Inference.} 
We utilize modality average weight without considering modality importance as a baseline \texttt{MoSE-AI}. For all modalities, we average the scores to obtain final prediction as Equation \eqref{eq:AI}.
\begin{equation}  \label{eq:AI}
    \mathcal{F}^{AI}(h,r,t) = \frac{1}{|\mathcal{M}|}\sum_{m \in \mathcal{M}}  f_m(h,r,t) \,.
\end{equation}

\textbf{Relation-aware Boosting Inference.}
We find that entity-relevant triples are sparse and thus hard to capture the accurate correlation between entity and modality importance.
In this paper, we assume that the relation of each modality varies in relevance level.
So we propose to learn modality weight in relation-level to adjust modality importance to final predictions.
We divide meta-set by relation type and upgrade RankBoost \cite{freund2003rankboost} mechanism to generate modality weights $w_m(r)$ corresponding to relation $r$ and combine modality scores as Equation \eqref{eq:BI}.
The \texttt{MoSE-BI} Algorithm is illustrated in Appendix \ref{app: boosting}.

\begin{equation} \label{eq:BI}
    \mathcal{F}^{BI}(h,r,t) = \sum_{m \in \mathcal{M}} w_m(r) f_m(h,r,t) \,.
\end{equation}

\textbf{Instance-specific Meta-Learner Inference.}
However, for KGs with fewer relations, such as WN9 \cite{xie2017IKRL} with only 9 relations, \texttt{MoSE-BI} is limited by coarse-grained relation-level weight learning.
Thus we propose to train a meta-learner to find optimal weight functions for each triple instance.
Following \newcite{shu2019metaweightnet}, we exploit an MLP (Multilayer Perceptron) with only one hidden layer as a meta-learner to combine the scores and approximate true predictions.
For a triple $(h,r,t)$, we use the concatenation of three scores $F(h,r,t) = [f_m(h,r,t)], m \in \mathcal{M}$ as input, and train the weighted scores to fit the final predictions. 
The final prediction is obtained as Equation \eqref{eq:MI} where weight parameter $\boldsymbol{\Theta}$ is trained in meta-set and transferred to test-set.
\begin{equation} \label{eq:MI}
    \mathcal{F}^{MI}((h,r,t);\boldsymbol{\Theta}) = w(\boldsymbol{\Theta}) F(h,r,t) \,.
\end{equation}

The optimal weight $\boldsymbol{\Theta}$ is obtained with CE loss as Equation \eqref{eq:ce meta}.
\begin{equation} \label{eq:ce meta}
    \mathcal{L}_{MI} = CE[softmax(\mathcal{F}^{MI}((h,r,t);\boldsymbol{\Theta}))] \,.
\end{equation}

\section{Experiments}

\subsection{Experimental Setting}
\begin{table}[!b]
\small
\centering
\resizebox{0.47\textwidth}{!}{
\begin{tabular}{crrrrr}
\toprule
Dataset   & \#Rel. & \#Ent. & \#Train & \#Valid & \#Test \\
\midrule
FB15K-237 & 237    & 14,541 & 272,115 & 17,535  & 20466  \\
WN18      & 18     & 40,943 & 141,442 & 5,000   & 5,000  \\
WN9       & 9      & 6,555  & 11,741  & 1,337   & 1,319 \\
\bottomrule
\end{tabular}
}
\caption{Datasets statistics for MKGC.}
\label{tab: datasets}
\end{table}
\begin{table*}[!t]
\centering
\resizebox{\linewidth}{!}{
\setlength{\tabcolsep}{1mm}{\begin{tabular}{lcccccccccccccc}

\toprule
\multirow{2}{*}{Model} & \multicolumn{4}{c}{FB15K-237} & \makebox[0.01\textwidth][c]{} & \multicolumn{4}{c}{WN18} &
\makebox[0.01\textwidth][c]{} &
\multicolumn{4}{c}{WN9} \\
\cline{2-5}
\cline{7-10}
\cline{12-15}
& Hits@1 $\uparrow$ & Hits@3 $\uparrow$ & Hits@10 $\uparrow$ & MR $\downarrow$ & & Hits@1 $\uparrow$ & Hits@3 $\uparrow$ & Hits@10 & $\uparrow$ MR $\downarrow$ & & Hits@1 $\uparrow$ & Hits@3 $\uparrow$ & Hits@10 $\uparrow$ & MR $\downarrow$ \\

\midrule
\multicolumn{10}{l}{\textit{Unimodal KGE methods}}\\
\midrule
TransE  & 0.198 & 0.376 & 0.441 & 323 && 0.040 & 0.745 & 0.923 & 357 && 0.864 & 0.901 & 0.917 & 146 \\
DistMult & 0.199 & 0.301 & 0.466 & 512 && 0.335 & 0.876 & 0.940 & 655 && 0.531 & 0.871 & 0.911 & 241 \\
ComplEx & 0.194 & 0.297 & 0.450 & 546 && 0.936 & 0.945 & 0.947 & -  && \underline{0.901} & 0.913 & 0.922 & 256 \\
RotatE  & 0.241 & 0.375 & 0.533 & 177 && 0.942 & 0.950 & 0.957 & 254 && 0.889 & 0.906 & 0.922 & 175 \\

\midrule
\multicolumn{10}{l}{\textit{Multimodal KGE methods}}\\
\midrule
IKRL (UNION) & 0.194 & 0.284 & 0.458 & 298 & & 0.127 & 0.796 & 0.928 & 596 & & - & - & 0.938 & 21 \\
TransAE & 0.199 & 0.317 & 0.463 & 431 & & 0.323 & 0.835 & 0.934 & 352 & & - & - & 0.942 & 17 \\
RSME  & 0.242 & 0.344 & 0.467 & 417 & & \underline{0.943} & 0.951 & 0.957 & 223 & & 0.878 & 0.912 & 0.923 & 55 \\
\midrule
\texttt{MoSE-AI} & 0.255 & 0.376 & 0.518 & 135 & & 0.929 & 0.946 & 0.962 & 23  & & 0.840	& \underline{0.932}	& 0.963 &	\textbf{4} \\
\texttt{MoSE-BI} & \textbf{0.281} & \textbf{0.411} & \textbf{0.565} & \textbf{117} & & 0.884 & \underline{0.953} & \underline{0.972} & \underline{8} & & 0.831 & 0.923 & \underline{0.964} & \textbf{4} \\
\texttt{MoSE-MI} &  \underline{0.268} & \underline{0.394} & \underline{0.540} & \underline{127} & & \textbf{0.948} & \textbf{0.962} & \textbf{0.974} & \textbf{7} & & \textbf{0.909} & \textbf{0.937} & \textbf{0.967} & \textbf{4} \\
\midrule
\midrule

\multicolumn{10}{l}{\textit{Pre-trained Language Model methods}}\\
\midrule
KG-BERT & - & - & 0.420 & 153 && 0.117 & 0.689 & 0.926 & 58 && - & - & - & - \\ 
MKGformer & 0.256 & 0.367 & 0.504 & 221 &  & 0.944 & 0.961 & 0.972 & 28 &  & - & - & - & - \\
\bottomrule
\end{tabular}}
}
\caption{Knowledge graph completion performance on FB15K-237, WN18, and WN9.
We highlight the \textbf{best} and the \underline{second best} results of each column. 
\texttt{MoSE-BI} performs the best on FB15k-237, and \texttt{MoSE-MI} achieves the best performance on WN18 and WN9. We can also observe that existing multimodal KGE methods do not perform as well as RotatE, which is a unimodal KGE method.
}
\label{tab: main results}
\end{table*}
\textbf{Datasets.}
To evaluate the proposed model, we conduct experiments on three widely used KGC datasets: FB15K-237 \cite{toutanova2015fb237}, WN18 \cite{bordes2013TransE}, and WN9-IMG \cite{xie2017IKRL}.
The former two are unimodal KGC datasets with only structure modal, and the latter one contains both structure and visual modalities.
We follow previous studies \cite{wang2021rsme, xie2016DKRL, yao2019kgbert} to augment the text and visual modality information of each dataset.
The dataset statistics are shown in Table \ref{tab: datasets}.

\textbf{Implementation details.}
To evaluate \texttt{MoSE}, four metrics are used, i.e., Hits@K, K=1, 3, 10, representing accuracy in top K predictions, and Mean Rank (MR).
Higher Hits@K and lower MR indicate better performance. 
We use Pytorch 1.11.0 to implement \texttt{MoSE}.
The operating system is Ubuntu 18.04.5.
We use a single NVIDIA A6000 GPU with 48GB of RAM.

We report the results of three \texttt{MoSE} variants which vary in the inference methods. \texttt{MoSE-AI} refers to \texttt{MoSE} with average inference. \texttt{MoSE-BI} refers to \texttt{MoSE} with boosting inference. \texttt{MoSE-MI} refers to \texttt{MoSE} with meta-learner inference.

We follow the widely-used filtered setting \cite{bordes2013TransE}, i.e., excluding other true entities when evaluating.
We exploit ComplEx \cite{lacroix2018fbkbc} as KGC Decoder.
In this paper, we mainly focus on contradiction in relation embeddings. 
Thus, we employ SOTA pretrained encoder to extract visual and text features of entities, i.e., ViT \cite{dosovitskiy2020vit} following RSME \cite{wang2021rsme} for visual modality and BERT \cite{kenton2019bert} for text modality. 
We use Adagrad \cite{duchi2011adagrad} to optimize the model.
The hyperparameters are selected with the best Hits@10 on the validation set.

\textbf{Baselines.}
We compare \texttt{MoSE} with several baselines to demonstrate the advantage of our framework. 
We mainly compare \texttt{MoSE} with KGE methods, which can be grouped into two categories:
(1) the unimodal KGE methods, including TransE \cite{bordes2013TransE}, DistMult \cite{yang2015DistMult}, ComplEx \cite{trouillon2016ComplEx}, RotatE \cite{sun2018RotatE}; 
(2) the multimodal KGE methods, including a) E-TCR methods: IKRL \cite{xie2017IKRL}, b) I-TCR methods: TransAE \cite{wang2019transAE} and RSME \cite{wang2021rsme}.
We also list the results of pre-trained language models (PLMs) for KGC, i.e., KG-BERT \cite{yao2019kgbert} and MKGformer \cite{chen2022mkgformer}.

\subsection{Comparison to the Baselines}
The experimental results in Table \ref{tab: main results} show that \texttt{MoSE} obtains the best performance compared to all 9 baselines, which demonstrates the superiority of \texttt{MoSE}.
Compared to unimodal KGE methods, \texttt{MoSE} outperforms the best unimodal method RotatE, while other multimodal methods do not.
Compared to multimodal KGE methods, \texttt{MoSE} achieves 2\% - 10\% improvements in Hits@10 and 13 - 216 improvements in MR over the best existing methods.
It is worth noting that even compared to the pre-trained language model methods,
\texttt{MoSE} outperforms KG-BERT and MKGformer in all metrics on FB15K-237 and WN18 datasets. 

\textit{Q1: Does \texttt{MoSE} succeed in avoiding modality interference?}
Compared with the corresponding base model, while other multimodal methods face a certain level of performance decline, \texttt{MoSE} achieves consistent improvements in all metrics.
For example, the Hits@1 and Hits@3 of IKRL drop compared to those of TransE, and the Hits@3 of TransAE drops compared to that of TransE on the FB15K-237 dataset.
Even SOTA RSME faces a slight drop on the WN9 dataset in terms of Hits@1 and Hits@3 compared to ComplEx.
It reveals that \texttt{MoSE} can steadily enhance unimodal models with auxiliary modality information and successfully avoid modality interference to structure modality.

\begin{table*}[!t]
\centering
\resizebox{\linewidth}{!}{
\setlength{\tabcolsep}{1mm}{\begin{tabular}{lcccccccccccccc}

\toprule
\multirow{2}{*}{Model} & \multicolumn{4}{c}{FB15K-237} & \makebox[0.01\textwidth][c]{} & \multicolumn{4}{c}{WN18} &
\makebox[0.01\textwidth][c]{} &
\multicolumn{4}{c}{WN9} \\
\cline{2-5}
\cline{7-10}
\cline{12-15}
& Hits@1 $\uparrow$ & Hits@3 $\uparrow$ & Hits@10 $\uparrow$ & MR $\downarrow$ & & Hits@1 $\uparrow$ & Hits@3 $\uparrow$ & Hits@10 & $\uparrow$ MR $\downarrow$ & & Hits@1 $\uparrow$ & Hits@3 $\uparrow$ & Hits@10 $\uparrow$ & MR $\downarrow$ \\
\midrule
\texttt{I-TCR}    & 0.192 & 0.303 & 0.439 & 439 &  & 0.945 & 0.953 & 0.958 & 298 &  & 0.588 & 0.755 & 0.847 & 126 \\
\midrule
\texttt{E-TCR-AI} & 0.248 & 0.367 & 0.511 & 140 &  & 0.910 & 0.945 & 0.960 & 27  &  & 0.779 & 0.916 & 0.958 & 5   \\
\texttt{MoSE-AI}  & 0.255 & 0.376 & 0.518 & 135 &  & 0.929 & 0.946 & 0.962 & 23  &  & 0.840 & 0.932 & 0.963 & \textbf{4}   \\
\midrule
\texttt{E-TCR-BI} & 0.271 & 0.402 & 0.554 & 121 &  & 0.858 & 0.945 & 0.968 & 10  &  & 0.756 & 0.914 & 0.959 & \textbf{4}   \\
\texttt{MoSE-BI}  & \textbf{0.281} & \textbf{0.411} & \textbf{0.565} & \textbf{117} &  & 0.884 & 0.953 & 0.972 & 8   &  & 0.831 & 0.923 & 0.964 & \textbf{4}   \\
\midrule
\texttt{E-TCR-MI} & 0.247 & 0.367 & 0.510 & 135 &  & 0.924 & 0.956 & 0.971 & 12  &  & 0.878 & 0.930 & 0.958 & 5   \\
\texttt{MoSE-MI}  & 0.268 & 0.394 & 0.540 & 127 &  & \textbf{0.948} & \textbf{0.962} & \textbf{0.974} & \textbf{7}   &  & \textbf{0.909} & \textbf{0.937} & \textbf{0.967} & \textbf{4}    \\
\bottomrule
\end{tabular}}
}
\caption{Effectiveness of relation decoupling. For the \texttt{I-TCR} variation, we fuse the multimodal entities and exploit a shared relation representation, which yields a unified prediction.
For the \texttt{E-TCR} variation, we replace the modality-split relation embeddings of \texttt{MoSE} with a shared relation embedding, which yields three  prediction scores as well.
}
\label{tab: relation variation}
\end{table*}

\textit{Q2: Is it necessary to assess modality importance?}
We explored three inference methods with modality importance in different aspects.
\texttt{MoSE-AI} treats each modality equally and does not consider modality importance at all, while \texttt{MoSE-BI} considers modality importance in relation-level and \texttt{MoSE-MI} in instance-level.
As shown in Table \ref{tab: main results}, \texttt{MoSE-BI} performs the best on FB15K-237 and \texttt{MoSE-MI} performs the best on WN18 and WN9.
All the best inference methods on the three datasets outperform \texttt{MoSE-AI}.
It demonstrates the necessity of assessing modality importance for MKGC.

\textit{Q3: How to choose the suitable inference methods?}
As we can observe, different inference methods expert in different KG characteristics.
The relation-aware inference \texttt{MoSE-BI} performs better in complex KGs with extensive relation types such as FB15K-237 (237 relations) and fails in KGs with fewer relation types such as WN18 and WN9 (18 and 9 relations respectively) while instance-specific inference \texttt{MoSE-MI} performs the opposite.
The possible reason is that the inference methods are with different capabilities to approximate the optimal combination of modalities. 
\texttt{MoSE-BI} is easy to scale to KGs with more relations and able to achieve relatively better performance.
Though \texttt{MoSE-MI} performs the best in two datasets, we believe that the single layer MLP may still limit the fitting capability of \texttt{MoSE-MI}.

\subsection{Effectiveness of Relation Decoupling}
Since the TCR baselines in Table \ref{tab: main results} vary in KGC decoder and modality types,
we further investigated different TCR variations of \texttt{MoSE} under the same setting to demonstrate the effectiveness of relation decoupling. The results are presented in Table \ref{tab: relation variation}.
For \texttt{I-TCR} and \texttt{E-TCR} variation, we replace the modality-split relation embeddings in \texttt{MoSE} with a single modality-shared relation embedding.
For \texttt{I-TCR} variation, we further fuse the multimodal entities with weighted concatenation, which yields a unified prediction.

As shown in Table \ref{tab: relation variation}, \texttt{MoSE} outperforms all its \texttt{E-TCR} variations under the same inference method.
As for \texttt{I-TCR} method, the best performance of \texttt{MoSE} exceeds \texttt{I-TCR} in all metrics. 
It demonstrates the necessity of modality relation decoupling.
We also notice that \texttt{I-TCR} exceeds \texttt{MoSE-AI} in Hits@1/3 and \texttt{MoSE-BI} in Hits@1 on WN18.
The possible reason is that the modality information of WN18 has many mutual semantics. So modality fusion brings accurate entity representations. 
However, \texttt{I-TCR} obtains a large MR score, indicating it is not stable as \texttt{MoSE} for MKGC. 
\begin{table*}[!t]
\centering
\resizebox{\linewidth}{!}{
\setlength{\tabcolsep}{1mm}{\begin{tabular}{lcccccccccccccc}

\toprule
\multirow{2}{*}{Model} & \multicolumn{4}{c}{FB15K-237} & \makebox[0.01\textwidth][c]{} & \multicolumn{4}{c}{WN18} &
\makebox[0.01\textwidth][c]{} &
\multicolumn{4}{c}{WN9} \\
\cline{2-5}
\cline{7-10}
\cline{12-15}
& Hits@1 $\uparrow$ & Hits@3 $\uparrow$ & Hits@10 $\uparrow$ & MR $\downarrow$ & & Hits@1 $\uparrow$ & Hits@3 $\uparrow$ & Hits@10 & $\uparrow$ MR $\downarrow$ & & Hits@1 $\uparrow$ & Hits@3 $\uparrow$ & Hits@10 $\uparrow$ & MR $\downarrow$ \\
\midrule
\texttt{MoSE-BEST}   & \textbf{0.281} & \textbf{0.411} & \textbf{0.565} & \textbf{117} && \textbf{0.948} & \textbf{0.962} & \textbf{0.974} & \textbf{7} & & \textbf{0.909} & \textbf{0.937} & \textbf{0.967} & \textbf{4} \\
\midrule
\texttt{Str-Str-Str-AI} & 0.256 & 0.386 & 0.542 & 166 &  & 0.945 & 0.954 & 0.960 & 247  &  & 0.909 & 0.916 & 0.923 & 201 \\
\texttt{Str-Str-Str-BI} & 0.262 & 0.392 & 0.547 & 162 &  & 0.946 & 0.954 & 0.960 & 247  &  & 0.909 & 0.916 & 0.923 & 200 \\
\texttt{Str-Str-Str-MI} & 0.256 & 0.386 & 0.541 & 206 &  & 0.945 & 0.954 & 0.960 & 322  &  & 0.909 & 0.916 & 0.923 & 263 \\
\midrule
\texttt{MoSE-Str}       & 0.264 & 0.392 & 0.545 & 168 &  & 0.946 & 0.954 & 0.960 & 264  &  & 0.908 & 0.914 & 0.922 & 193 \\
\texttt{MoSE-Vis}       & 0.167 & 0.242 & 0.329 & 890 &  & 0.527 & 0.611 & 0.685 & 2017 &  & 0.092 & 0.231 & 0.392 & 243 \\
\texttt{MoSE-Text}      & 0.245 & 0.364 & 0.500 & 161 &  & 0.255 & 0.442 & 0.618 & 96   &  & 0.262 & 0.487 & 0.709 & 27  \\

\bottomrule
\end{tabular}}
}
\caption{The experimental results of modality ablation. 
\texttt{MoSE-BEST} refers to results obtained by the best variant of \texttt{MoSE} for each dataset, i.e., \texttt{MoSE-BI} for FB15K-237 and \texttt{MoSE-MI} for WN18 and WN9.
\texttt{Str}, \texttt{Vis}, \texttt{Text} refer to structure, visual and text modality respectively. 
\texttt{Str-Str-Str-AI/BI/MI} refers to results obtained by replacing both visual and text modalities in \texttt{MoSE} with structure modality.
\texttt{MoSE-Str/Vis/Text} refers to the modality-split prediction performances of each modality.
}
\label{tab: ablation results}
\end{table*}

\subsection{Modality Ablation}
To demonstrate how each modality supports final predictions, we conduct modality ablation.
Table \ref{tab: ablation results} shows the experimental results obtained by 
(1) ensemble inference of three structure unimodal models \texttt{Str-Str-Str-AI/BI/MI}, 
(2) modality-split predictions obtained by KGC decoder \texttt{Mose-Str/Vis/Text}.

The improvements of \texttt{Str-Str-Str} over \texttt{MoSE-Str} is insignificant compared to that of \texttt{MoSE-BEST} over \texttt{MoSE-Str}.
It reveals that \texttt{MoSE} improves the base unimodal model via effectively utilizing modality information instead of performing ensemble inference.
For modality-split predictions \texttt{MoSE-Str/Vis/Text}, no single one of three prediction performances exceeds \texttt{MoSE-BEST}.
It demonstrates that modalities in \texttt{MoSE} effectively enhance each other and successfully avoid modality mutual interference.
The modality-split predictions also indicate the modality quality for assisting MKGC.
The structure modality, which is directly learned from KGs, remains the best performance on all datasets, while visual modality has erratic performance and text modality consistently provides the best MR metric.

\subsection{Case Study}
\begin{figure}[t]

     \centering
     \begin{subfigure}[b]{0.47\textwidth}
         \centering
         \includegraphics[width=\textwidth]{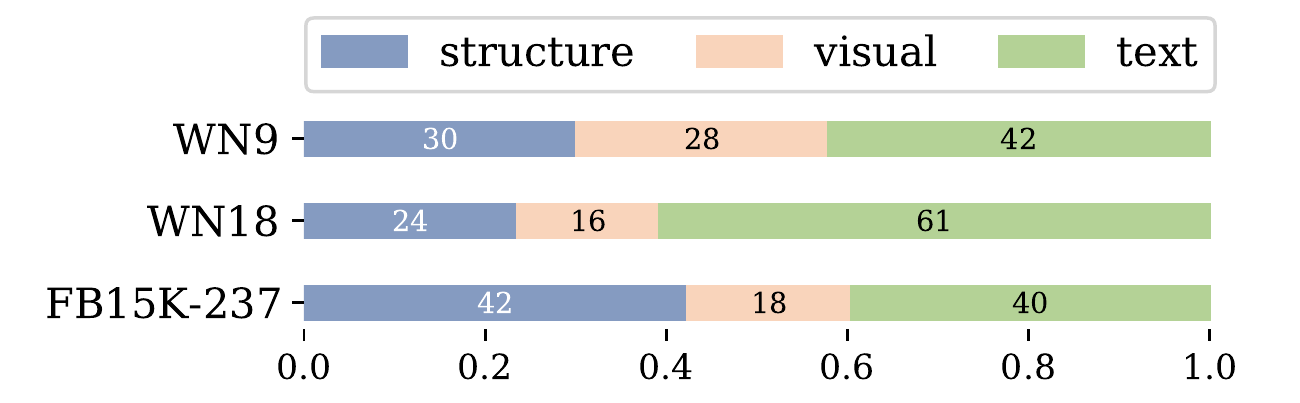}
         \caption{Average modality weights (\%) of datasets.}
         \label{fig:case_dataset}
     \end{subfigure}
    \vspace*{\fill}
     \begin{subfigure}[b]{0.47\textwidth}
         \centering
         \includegraphics[width=\textwidth]{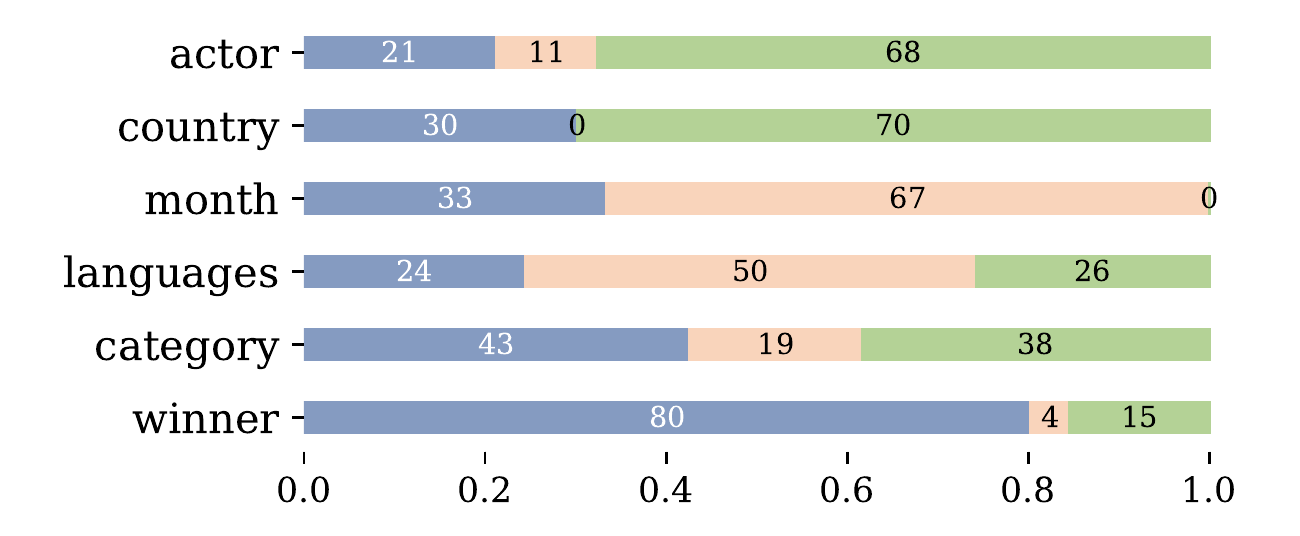}
         \caption{Examples of modality weights (\%) of FB15K-237 relations. Relations are abbreviated. See Appendix \ref{app: rel} for full relation names.}
         \label{fig:case_rels}
     \end{subfigure}
     \caption{Modality weights (\%) learned by \texttt{MoSE-BI}. }
     \label{fig: case}
\end{figure}

To demonstrate the intuitive ability of \texttt{MoSE} to assess modality importance, we conduct case studies with \texttt{MoSE-BI}, which provides modality weights for each modality corresponding to each relation. 
Figure \ref{fig: case} shows modality weights in Equation \eqref{eq:BI} to combine predictions from multiple sources.

\textbf{Modality Importance.}
Figure \ref{fig:case_dataset} shows average modality weights on each dataset obtained by \texttt{MoSE-BI}. 
We can observe that text modality provides the greatest contributions on WN18 and WN9, while visual modality provides the minimum on all datasets.
It demonstrates that text modality provides valuable information supporting knowledge predictions while visual modality in the opposite.
The possible reason is that descriptions often mention relevant entities, while images are only highly related to entity itself.

\textbf{Relation Cases.}
Figure \ref{fig:case_rels} presents some examples to show how much each modality contributes to relation learning on FB15K-237.
The higher level of modality importance often stems from more relation-relevant modality information. 
For example, for relation \texttt{country\_of\_origin} (abbr. \texttt{country}) shown in Figure \ref{fig:intro_example}, the text modality provides more relevance information than visual modality. As shown in Figure \ref{fig:case_rels}, text modality presents importance up to 70\% while visual modality presents 0\%.
The results also demonstrate that \texttt{MoSE-BI} is able to identify which modality is more credible and then assign a higher weight in a fine-grained relation level.

\begin{figure}[!t]
    \centering
    \includegraphics[width=0.48\textwidth]{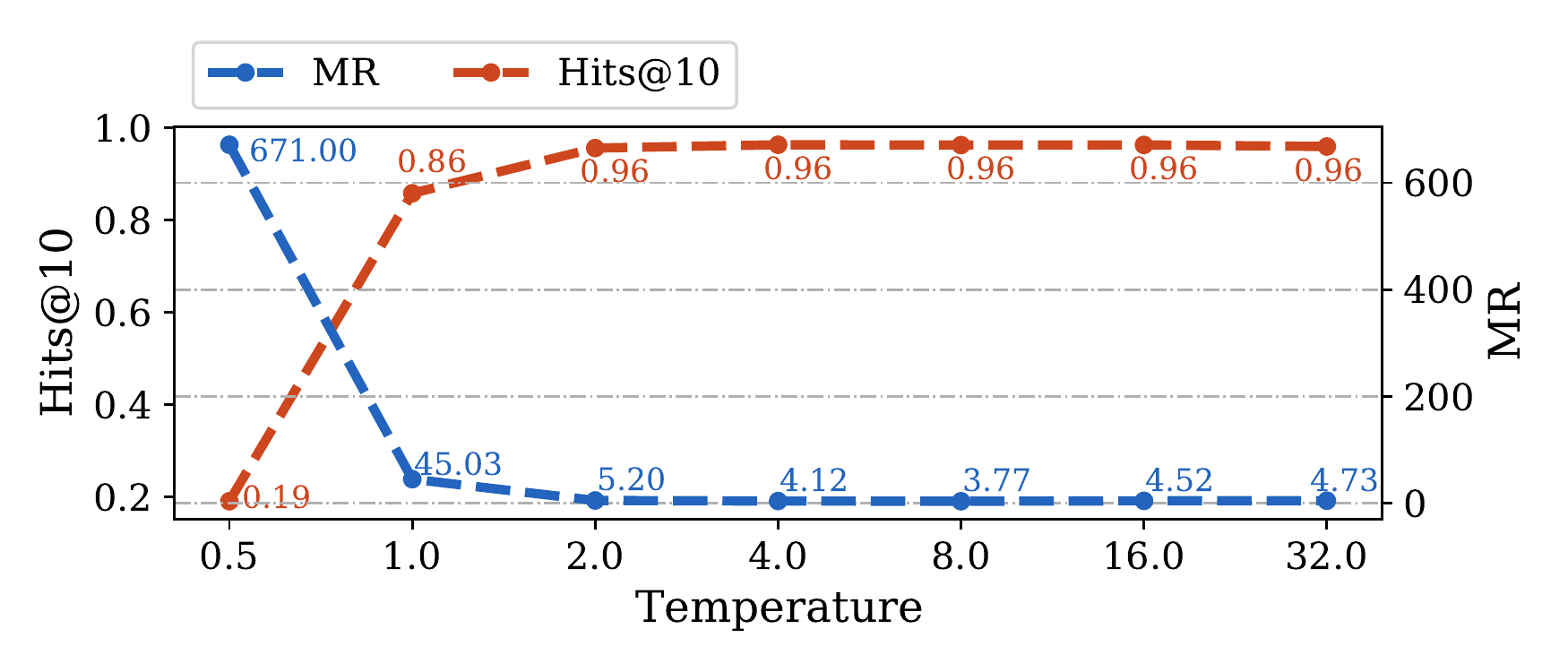}
    \caption{Temperature Parameter Analysis conducted by MoSE-AI on WN9.}
    \label{fig: temp}
\end{figure}

\subsection{Uncertainty in MKGs}
To investigate the uncertainty of MKG predictions, we adjust the temperature parameter as shown in Figure \ref{fig: temp}.
We use \texttt{MoSE-AI} to rule out the impact of ensemble inference.
We vary the temperature $\mathcal{T}$ in Equation \eqref{eq: cc} from
$2^{-1}$ to $2^5$ with exponential growth. 
As the temperature increases, the performance tends to grow and converge to stable.
When $\mathcal{T}=2^{-1}$, the confidence to visual and text modality is enlarged and \texttt{MoSE} faces great performance decline.
We can also observe that \texttt{MoSE} with larger $\mathcal{T}$ always outperforms $\mathcal{T}=2^0=1$ in which the confidence is not constrained.
It proves our assumption about the uncertainty of visual and text modalities.

\section{Conclusion}
In this paper, we propose a novel modality split learning and ensemble inference framework for multimodal knowledge graph completion called \texttt{MoSE}.
\texttt{MoSE} first decouples modality-shared relation embedding to modality-split relation embeddings and performs modality-split representation learning in the training phase, aiming at overcoming modality relation contradiction.
Then, \texttt{MoSE} exploits three ensemble inference techniques to combine the modality-split predictions by assessing modality importance.
Experiment results demonstrate that \texttt{MoSE} outperforms state-of-the-art methods for MKGC task on three widely-used datasets.


\section*{Limitations}
Despite that \texttt{MoSE} achieves some gains by modality-split learning and ensemble inference, \texttt{MoSE} still has the following limitations:

First, \texttt{MoSE} does not fully exploit visual modality.
Since the image of the entity is highly \textit{self-relevant} and covers less information about \textit{other} related entities, we reduce the visual modality importance during ensemble inference to cater to the MKGC task, which heavily relies on the relationship between entities. Nevertheless, we believe there are other ways to exploit visual modality suitably.

Second, for a fair comparison, we follow SOTA method RSME \cite{wang2021rsme} and utilize a single-image setting. We believe that under the multiple-image setting, the problem of modality relation contradiction still holds. 
Intuitively, even with more images, the image of the entity "The United States of America" in Figure \ref{fig:intro_example} is unlikely to involve the entity "Friends".
Quantitatively, the similarity of multiple images from the same entity is up to 99.250\% on FB15K-237 and 99.255\% on WN18 respectively. 
Therefore, there is little difference between single-image and multiple-image settings in our work.
However, more images may introduce more side information, such as related entities, from which MKGC model may benefit.

\section*{Acknowledgements}
This research is supported by the National Natural Science Foundation of China
(No. U1936206, 62272250, 62002178, U1903128).

\bibliography{ref_simplified,custom}

\begin{thebibliography}{24}
\expandafter\ifx\csname natexlab\endcsname\relax\def\natexlab#1{#1}\fi

\bibitem[{Bordes et~al.(2013)Bordes, Usunier, Garcia-Duran, Weston, and
  Yakhnenko}]{bordes2013TransE}
Antoine Bordes, Nicolas Usunier, Alberto Garcia-Duran, Jason Weston, and Oksana
  Yakhnenko. 2013.
\newblock Translating embeddings for modeling multi-relational data.
\newblock \emph{Advances in neural information processing systems}, 26.

\bibitem[{Chen et~al.(2022)Chen, Zhang, Li, Deng, Tan, Xu, Huang, Si, and
  Chen}]{chen2022mkgformer}
Xiang Chen, Ningyu Zhang, Lei Li, Shumin Deng, Chuanqi Tan, Changliang Xu, Fei
  Huang, Luo Si, and Huajun Chen. 2022.
\newblock Hybrid transformer with multi-level fusion for multimodal knowledge
  graph completion.
\newblock \emph{arXiv preprint arXiv:2205.02357}.

\bibitem[{Chen et~al.(2020)Chen, Chen, Fan, Uppunda, Sun, and
  Zaniolo}]{chen2020multilingual}
Xuelu Chen, Muhao Chen, Changjun Fan, Ankith Uppunda, Yizhou Sun, and Carlo
  Zaniolo. 2020.
\newblock Multilingual knowledge graph completion via ensemble knowledge
  transfer.
\newblock In \emph{Findings of the Association for Computational Linguistics:
  EMNLP 2020}, pages 3227--3238.

\bibitem[{Dosovitskiy et~al.(2020)Dosovitskiy, Beyer, Kolesnikov, Weissenborn,
  Zhai, Unterthiner, Dehghani, Minderer, Heigold, Gelly
  et~al.}]{dosovitskiy2020vit}
Alexey Dosovitskiy, Lucas Beyer, Alexander Kolesnikov, Dirk Weissenborn,
  Xiaohua Zhai, Thomas Unterthiner, Mostafa Dehghani, Matthias Minderer, Georg
  Heigold, Sylvain Gelly, et~al. 2020.
\newblock An image is worth 16x16 words: Transformers for image recognition at
  scale.
\newblock In \emph{ICLR}.

\bibitem[{Duchi et~al.(2011)Duchi, Hazan, and Singer}]{duchi2011adagrad}
John Duchi, Elad Hazan, and Yoram Singer. 2011.
\newblock Adaptive subgradient methods for online learning and stochastic
  optimization.
\newblock \emph{Journal of machine learning research}, 12(7).

\bibitem[{Freund et~al.(2003)Freund, Iyer, Schapire, and
  Singer}]{freund2003rankboost}
Yoav Freund, Raj Iyer, Robert~E Schapire, and Yoram Singer. 2003.
\newblock An efficient boosting algorithm for combining preferences.
\newblock \emph{Journal of machine learning research}, 4(Nov):933--969.

\bibitem[{Guo et~al.(2017)Guo, Pleiss, Sun, and
  Weinberger}]{guo2017tempscaling}
Chuan Guo, Geoff Pleiss, Yu~Sun, and Kilian~Q Weinberger. 2017.
\newblock On calibration of modern neural networks.
\newblock In \emph{ICML}, pages 1321--1330. PMLR.

\bibitem[{Kenton and Toutanova(2019)}]{kenton2019bert}
Jacob Devlin Ming-Wei~Chang Kenton and Lee~Kristina Toutanova. 2019.
\newblock Bert: Pre-training of deep bidirectional transformers for language
  understanding.
\newblock In \emph{Proceedings of NAACL-HLT}, pages 4171--4186.

\bibitem[{Lacroix et~al.(2018)Lacroix, Usunier, and
  Obozinski}]{lacroix2018fbkbc}
Timoth{\'e}e Lacroix, Nicolas Usunier, and Guillaume Obozinski. 2018.
\newblock Canonical tensor decomposition for knowledge base completion.
\newblock In \emph{ICML}, pages 2863--2872. PMLR.

\bibitem[{Marino et~al.(2019)Marino, Rastegari, Farhadi, and
  Mottaghi}]{marino2019okvqa}
Kenneth Marino, Mohammad Rastegari, Ali Farhadi, and Roozbeh Mottaghi. 2019.
\newblock Ok-vqa: A visual question answering benchmark requiring external
  knowledge.
\newblock In \emph{CVPR}, pages 3195--3204.

\bibitem[{Mousselly-Sergieh et~al.(2018)Mousselly-Sergieh, Botschen, Gurevych,
  and Roth}]{mousselly2018MKB}
Hatem Mousselly-Sergieh, Teresa Botschen, Iryna Gurevych, and Stefan Roth.
  2018.
\newblock A multimodal translation-based approach for knowledge graph
  representation learning.
\newblock In \emph{Proceedings of the Seventh Joint Conference on Lexical and
  Computational Semantics}, pages 225--234.

\bibitem[{Shu et~al.(2019)Shu, Xie, Yi, Zhao, Zhou, Xu, and
  Meng}]{shu2019metaweightnet}
Jun Shu, Qi~Xie, Lixuan Yi, Qian Zhao, Sanping Zhou, Zongben Xu, and Deyu Meng.
  2019.
\newblock Meta-weight-net: Learning an explicit mapping for sample weighting.
\newblock \emph{Advances in neural information processing systems}, 32.

\bibitem[{Sun et~al.(2020)Sun, Cao, Zhao, Wan, Zhou, Zhang, Wang, and
  Zheng}]{sun2020mmkgrs}
Rui Sun, Xuezhi Cao, Yan Zhao, Junchen Wan, Kun Zhou, Fuzheng Zhang, Zhongyuan
  Wang, and Kai Zheng. 2020.
\newblock Multi-modal knowledge graphs for recommender systems.
\newblock In \emph{CIKM}, pages 1405--1414.

\bibitem[{Sun et~al.(2018)Sun, Deng, Nie, and Tang}]{sun2018RotatE}
Zhiqing Sun, Zhi-Hong Deng, Jian-Yun Nie, and Jian Tang. 2018.
\newblock Rotate: Knowledge graph embedding by relational rotation in complex
  space.
\newblock In \emph{ICLR}.

\bibitem[{Toutanova et~al.(2015)Toutanova, Chen, Pantel, Poon, Choudhury, and
  Gamon}]{toutanova2015fb237}
Kristina Toutanova, Danqi Chen, Patrick Pantel, Hoifung Poon, Pallavi
  Choudhury, and Michael Gamon. 2015.
\newblock Representing text for joint embedding of text and knowledge bases.
\newblock In \emph{EMNLP}, pages 1499--1509.

\bibitem[{Trouillon et~al.(2016)Trouillon, Welbl, Riedel, Gaussier, and
  Bouchard}]{trouillon2016ComplEx}
Th{\'e}o Trouillon, Johannes Welbl, Sebastian Riedel, {\'E}ric Gaussier, and
  Guillaume Bouchard. 2016.
\newblock Complex embeddings for simple link prediction.
\newblock In \emph{ICML}.

\bibitem[{Wang et~al.(2021)Wang, Wang, Yang, Zhang, Chen, and
  Qi}]{wang2021rsme}
Meng Wang, Sen Wang, Han Yang, Zheng Zhang, Xi~Chen, and Guilin Qi. 2021.
\newblock Is visual context really helpful for knowledge graph? a
  representation learning perspective.
\newblock In \emph{Proceedings of the 29th ACM International Conference on
  Multimedia}, pages 2735--2743.

\bibitem[{Wang et~al.(2019)Wang, Li, Li, and Zeng}]{wang2019transAE}
Zikang Wang, Linjing Li, Qiudan Li, and Daniel Zeng. 2019.
\newblock Multimodal data enhanced representation learning for knowledge
  graphs.
\newblock In \emph{2019 International Joint Conference on Neural Networks
  (IJCNN)}, pages 1--8. IEEE.

\bibitem[{Xie et~al.(2016)Xie, Liu, Jia, Luan, and Sun}]{xie2016DKRL}
Ruobing Xie, Zhiyuan Liu, Jia Jia, Huanbo Luan, and Maosong Sun. 2016.
\newblock Representation learning of knowledge graphs with entity descriptions.
\newblock In \emph{AAAI}, volume~30.

\bibitem[{Xie et~al.(2017)Xie, Liu, Luan, and Sun}]{xie2017IKRL}
Ruobing Xie, Zhiyuan Liu, Huanbo Luan, and Maosong Sun. 2017.
\newblock Image-embodied knowledge representation learning.
\newblock In \emph{IJCAI}, pages 3140--3146.

\bibitem[{Yang et~al.(2015)Yang, Yih, He, Gao, and Deng}]{yang2015DistMult}
Bishan Yang, Scott Wen-tau Yih, Xiaodong He, Jianfeng Gao, and Li~Deng. 2015.
\newblock Embedding entities and relations for learning and inference in
  knowledge bases.
\newblock In \emph{ICLR}.

\bibitem[{Yao et~al.(2019)Yao, Mao, and Luo}]{yao2019kgbert}
Liang Yao, Chengsheng Mao, and Yuan Luo. 2019.
\newblock Kg-bert: Bert for knowledge graph completion.
\newblock \emph{arXiv preprint arXiv:1909.03193}.

\bibitem[{Zhang et~al.(2018)Zhang, Fu, Liu, and Huang}]{zhang2018mner}
Qi~Zhang, Jinlan Fu, Xiaoyu Liu, and Xuanjing Huang. 2018.
\newblock Adaptive co-attention network for named entity recognition in tweets.
\newblock In \emph{AAAI}.

\bibitem[{Zhu et~al.(2022)Zhu, Li, Wang, Jiang, Sun, Wang, Xiao, and
  Yuan}]{zhu2022mmkgsurvey}
Xiangru Zhu, Zhixu Li, Xiaodan Wang, Xueyao Jiang, Penglei Sun, Xuwu Wang,
  Yanghua Xiao, and Nicholas~Jing Yuan. 2022.
\newblock Multi-modal knowledge graph construction and application: A survey.
\newblock \emph{arXiv preprint arXiv:2202.05786}.

\end{thebibliography}
\bibliographystyle{acl_natbib}

\newpage
\section*{Appendix}

\appendix


\section{\texttt{MoSE} Boosting Inference Algorithm} \label{app: boosting}
In this section, we present the algorithm detail of relation-aware boosting inference method \texttt{MoSE-BI} in Algorithm \ref{alg:mosebi}. 
We first divide the meta-set $\mathcal{D}$ to relation-aware sets $\mathcal{D}_r$ by relation type.
At each set, we exploit RankBoost \cite{freund2003rankboost} algorithm to model modality importance and combine modality scores to obtain final predictions.

The main idea of RankBoost is to turn a ranking problem into a classification problem. 
The score of corrupted entity $h_m(h,r,e)$ less than that of true tail entity $h_m(h,r,t)$ is seen as True prediction while False prediction in the opposite. It is worth noting that we select the best modality in each round to eliminate the impact of modality order.

\begin{algorithm}[!b]
\small
	\caption{\texttt{MoSE} Boosting Inference Algorithm}
	\label{alg:mosebi}
	\begin{algorithmic}[1]
    \REQUIRE The meta-set $\mathcal{D} = \{(h,r,t)\}, h,t \in \mathcal{E},r\in \mathcal{R}$; Modality set $\mathcal{M}$; Modality-split scores $f_m(h,r,t), m\in\mathcal{M}$; 
	\ENSURE $\mathcal{F}^{BI}(h,r,t)=\sum_{m\in\mathcal{M}} w_m(r)f_m(h,r,t)$
	\STATE divide meta-set $\mathcal{D}$ by relation $r \in \mathcal{R}$ to $\mathcal{D} = \{\mathcal{D}_r\}$
	\FOR{each relation set $\mathcal{D}_r \in \mathcal{D}$}
	    \STATE init $D_1(e) = \frac{1}{|\mathcal{M}|}$, $w_m(r) = 0, m\in\mathcal{M}$ \\
        \FOR{each modality $m=1, ..., |\mathcal{M}|$}
            \STATE $h_m(e) = \begin{cases} 1, f_m(h,r,e)<f_m(h,r,t) \\ -1, f_m(h,r,e)>=f_m(h,r,t) \end{cases}$
            \FOR{$i = 1, ..., |\mathcal{M}|$}
                \STATE $w_m^i = 
    \dfrac{1}{2}ln(\dfrac{\sum_{e\in\mathcal{E},h_m(e)=1} D_m(e)}{\sum_{e\in\mathcal{E},h_m(e)=-1} D_m(e)})$ \\
            \ENDFOR
            \STATE select the best and unchosen modality with max weight $w_m = \max\{w_m^1, ..., w_m^{|\mathcal{M}|}\}$
            \STATE $Z_m = \sum_{e\in \mathcal{E}} D_m(e) exp(-w_m h_m(e))$
            \STATE $D_{m+1}(e) = \dfrac{D_m(e) exp(-w_m h_m(e))}{Z_m}$ \\
            \STATE $w_m(r) = w_m(r) + w_m$ \\
        \ENDFOR
    \ENDFOR
	\RETURN $\mathcal{F}^{BI}(h,r,t)=\sum_{m\in\mathcal{M}}w_m(r)f_m(h,r,t)$
    \end{algorithmic}
\end{algorithm}

\section{Relation names} \label{app: rel}
Table \ref{tab: relnames} shows original relation names in Figure \ref{fig:case_rels}.

\begin{table}[!b]
\small
\centering
\resizebox{0.5\textwidth}{!}{
\begin{tabular}{cc}
\toprule
relation abbv.  & relation name \\
\midrule
\texttt{actor}     & \makecell[c]{/tv/tv\_program/regular\_cast./tv/\\regular\_tv\_appearance/actor}\\
\midrule
\texttt{country}   & \makecell[c]{/tv/tv\_program/country\_of\_origin} \\
\midrule
\texttt{month}     & \makecell[c]{/travel/travel\_destination/climate./\\travel/travel\_destination\_monthly\_climate/month} \\
\midrule
\texttt{languages} & \makecell[c]{/tv/tv\_program/languages} \\
\midrule
\texttt{category}  & \makecell[c]{/award/award\_category/category\_of} \\
\midrule
\texttt{winner}    & \makecell[c]{/award/award\_ceremony/awards\_presented./\\award/award\_honor/award\_winner} \\
\bottomrule
\end{tabular}
}
\caption{Relation abbreviations and full names.}
\label{tab: relnames}
\end{table}

\end{document}